\pdfoutput=1
\documentclass[11pt]{article}

\usepackage[final]{acl}

\usepackage{times}
\usepackage{latexsym}
\usepackage{amssymb} % For \mathbb
\usepackage{amsmath}

\usepackage[T1]{fontenc}
\usepackage[utf8]{inputenc}

\usepackage{microtype}

\usepackage{inconsolata}

\usepackage{graphicx}

\title{Quantifying Symptom Causality in Clinical Decision Making: \\ An Exploration Using CausaLM}

\author{
  Alexander Romanus \\
  \texttt{romanus@usc.edu} \\\And
  Connor Jordan \\
  \texttt{cjordan9@usc.edu} \\\And
  Mehul Shetty \\
  \texttt{mehulshe@usc.edu}}

\begin{document}
\maketitle
\begin{abstract}
Current machine learning approaches to medical diagnosis often rely on correlational patterns between symptoms and diseases, risking misdiagnoses when symptoms are ambiguous or common across multiple conditions. In this work, we move beyond correlation to investigate the causal influence of key symptoms—specifically “chest pain”—on diagnostic predictions. Leveraging the \textit{CausaLM} framework, we generate counterfactual text representations in which target concepts are effectively “forgotten,” enabling a principled estimation of the causal effect of that concept on a model’s predicted disease distribution. By employing Textual Representation-based Average Treatment Effect (TReATE), we quantify how the presence or absence of a symptom shapes the model’s diagnostic outcomes, and contrast these findings against correlation-based baselines such as CONEXP. Our results offer deeper insight into the decision-making behavior of clinical NLP models and have the potential to inform more trustworthy, interpretable, and causally-grounded decision support tools in medical practice.\end{abstract}

\section{Introduction}

One of the key issues in causal inference in NLP is the generation of counterfactuals. To generate counterfactuals we need a controlled setting where it is possible to compute the differences between actual text and what the text would have been if a specific concept in the text had not existed. Here, a concept refers to an entire semantic space. While there are many methods to generate counterfactuals manually, they are often arduous and impossible to generate for large datasets.

It is much easier to instead generate a counterfactual text representation based on adversarial learning than to create counterfactual text. We plan to use the \textit{CausaLM} \citep{10.1162/coli_a_00404} framework to create counterfactual text representations for clinical notes that "forget" a target concept while making sure it does not forget other potential confounders called control concepts. This innovative approach leverages the power of causal language models to uncover the intricate relationships between symptoms and diagnoses in medical decision-making. Specifically, we will use medical reports describing patient symptoms that is ingestible by \textit{CausaLM} through tokenization. Once the data is processed in a structured format, it will be optimal for use in the \textit{CausaLM} model. We will then use it to predict the counterfactual diagnoses as if certain symptoms had not been considered by the doctor---allowing us to explore alternative diagnostic pathways, and the weights given to various symptoms in the diagnostic process.

By implementing this approach, we can simulate situations where these symptoms were either not observed or not reported, providing insight into how the absence of key symptoms might alter the diagnostic outcome. We hope this will give us an accurate measure of the causal effect that these symptoms have on a specialist referral, creating a more robust and systematic way to compare diagnoses. This research could have far-reaching implications for improving clinical decision support systems, enhancing medical education, and ultimately improving patient outcomes through more precise and personalized diagnostic approaches.

\section{Related Work}

Making inferences about patients' diagnoses based on symptoms is commonly seen in many NLP models in medical contexts. 

Research has focused on finding the relationship between clinical notes and the doctors' resulting diagnosis using LLM. \citet{mullenbach-etal-2018-explainable} show NLP models making predictions for resulting ICD codes from patient data. 

\citet{toma-etal-2024-wanglab-mediqa} combine both text and images as input to provide medical queries using LLMs. \citet{michalopoulos-etal-2021-umlsbert} combine the uses of common NLPs with their own context embedding system to understand and make accurate predictions based on large amounts of clinical text. While these models make accurate diagnoses, none of them apply causality to their predictions as we proposed.

Causal machine learning concepts have also been used in the medical setting to improve patient outcomes, such as \citet{Richens2020} where causal machine learning is used as a way to improve the accuracy of the various algorithms and frameworks in the model to properly diagnoses rare diseases. The LLM proposed in this project differs with its use of causal reasoning compared to the causal machine learning concepts used by \citet{Richens2020}.

\citet{gopalakrishnan2024causalityextractionmedicaltext} used automated techniques to identify causalities from an annotated set of medical data. The researchers in this study utilize BioBERT to perform causal extraction tasks, enabling the collection of large datasets to enhance clinical decision-making and patient care.

\citet{ganin2016domainadversarialtrainingneuralnetworks} utilizes a technique called a "gradient reversal layer" to adversarially learn domain-invariant features. In their approach, they were able to successfully improve performance on various tasks such as sentiment analysis and image classification through domain invariance. In our work, we can use the concept of the gradient reversal layer to help our model "forget" a symptom and in turn adversarially train it and measure the impact of different symptoms.

Perhaps the work that we derive most heavily from is \citet{10.1162/coli_a_00404}. We will be using the underling framework proposed in this paper but finetuning it specifically for clinical decision making.

\section{Methodology}

To describe our methodology we first need to define causal model explanation. While causal inference is the main objective in many scientific endeavors, we rely here on a completely different aspect called causal model explanation. We're trying to find the causal effect of a given concept (called treatment) on the model's predictions, and show these effects to explain the observed behavior of the model. Here, a concept refers to a higher level, often aggregated unit, atomic input features such as words. It's an abstract idea rather that a collection of words.

\subsection{Language Representation}

Our approach is based on the idea that any text is created by a number of concepts coming together through a data-generating process \citep{10.1162/coli_a_00404}. Figure 1 describes this process. Imagine that we observe a clinical note $X$ and have trained a model to give a disease distribution based on the symptoms. We can hypothesize a list of concepts that might affect the model's decisions. We will denote the set of binary variables $C = \{C_j \in \{0,1\} | j \in \{0,1,2,...,k\}$, where each variable corresponds to the existence of a predefined concept (in our case each symptom). If $C_j = 1$ that means the j-th symptom exists in the text. We assume a pretrained language model $\phi$ and wish to assert how our trained disease distribution model $f$ is affected by the concepts in $C$. 

\begin{figure}[h]
    \centering
    \includegraphics[width=0.5\textwidth]{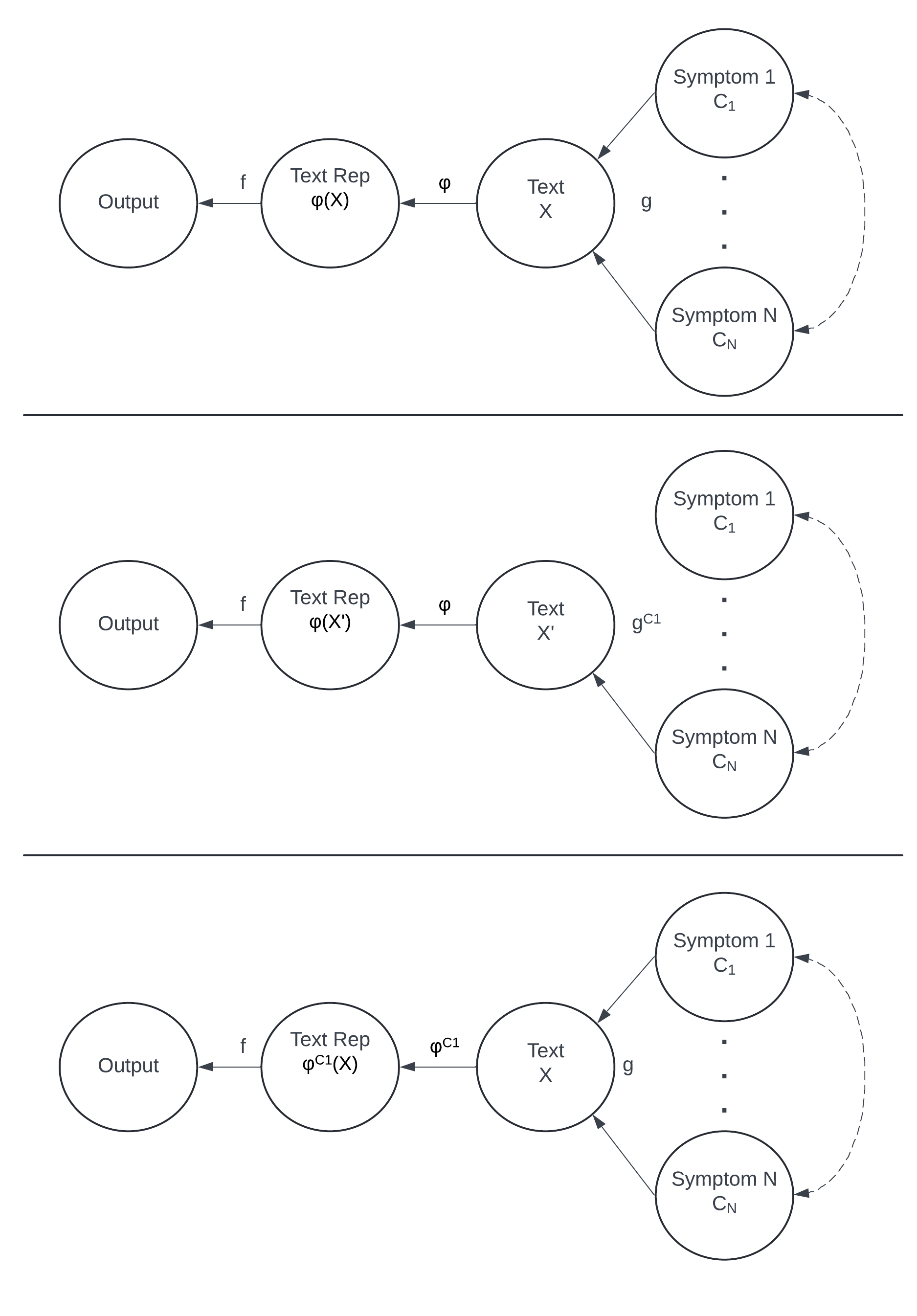}
    \caption{Three causal graphs relating the concepts of symptoms, clinical notes, their representations and distribution output.}
\end{figure}

The top graph describes the original data-generating process $g$. The middle graph describes the case of directly manipulating the text using an alternative generative process $g^{C1}$ that allows us to create a text representation $X'$ that is the same as $X$ but does not contain the concept $C_1$. The bottom graph describes our approach where we manipulate the text representation and not the text itself. The dashed edges between concepts indicate possible hidden confounders.

\subsection{Representation-Based Counterfactual Generation}

Next we will discuss how went about creating our model using the \textit{CausaLM} framework. We first used a pre-trained BERT and finetuned it for our goal. As described in Figure 2, we used the original BERT NSP and MLM heads and added a new TC head that had the task of determining if the treatment concept exists in the text or not. We added a cross entropy loss for the TC classification task to the MLM and NSP loss.

\begin{center}
    $Loss_{total} = Loss_{MLM} + Loss_{NSP} + Loss_{CE} $
\end{center}

You can add an additional CC head to determine if the control concept exists in the text but we did not add that in our project to maintain simplicity. We then added a gradient reversal layer between the TC head and the rest of the model to reverse the gradients of the loss for the TC classification task by a factor $\lambda$. We continued learning on the NSP and MLM tasks because we wanted our model to stay just as good at those tasks as it was before we finetuned it. Our final goal was to maximize the TC classification loss and minimize the loss for the MLM and NSP tasks. This is the BERT-TC model.

\begin{figure}[h]
    \centering
    \includegraphics[width=0.5\textwidth]{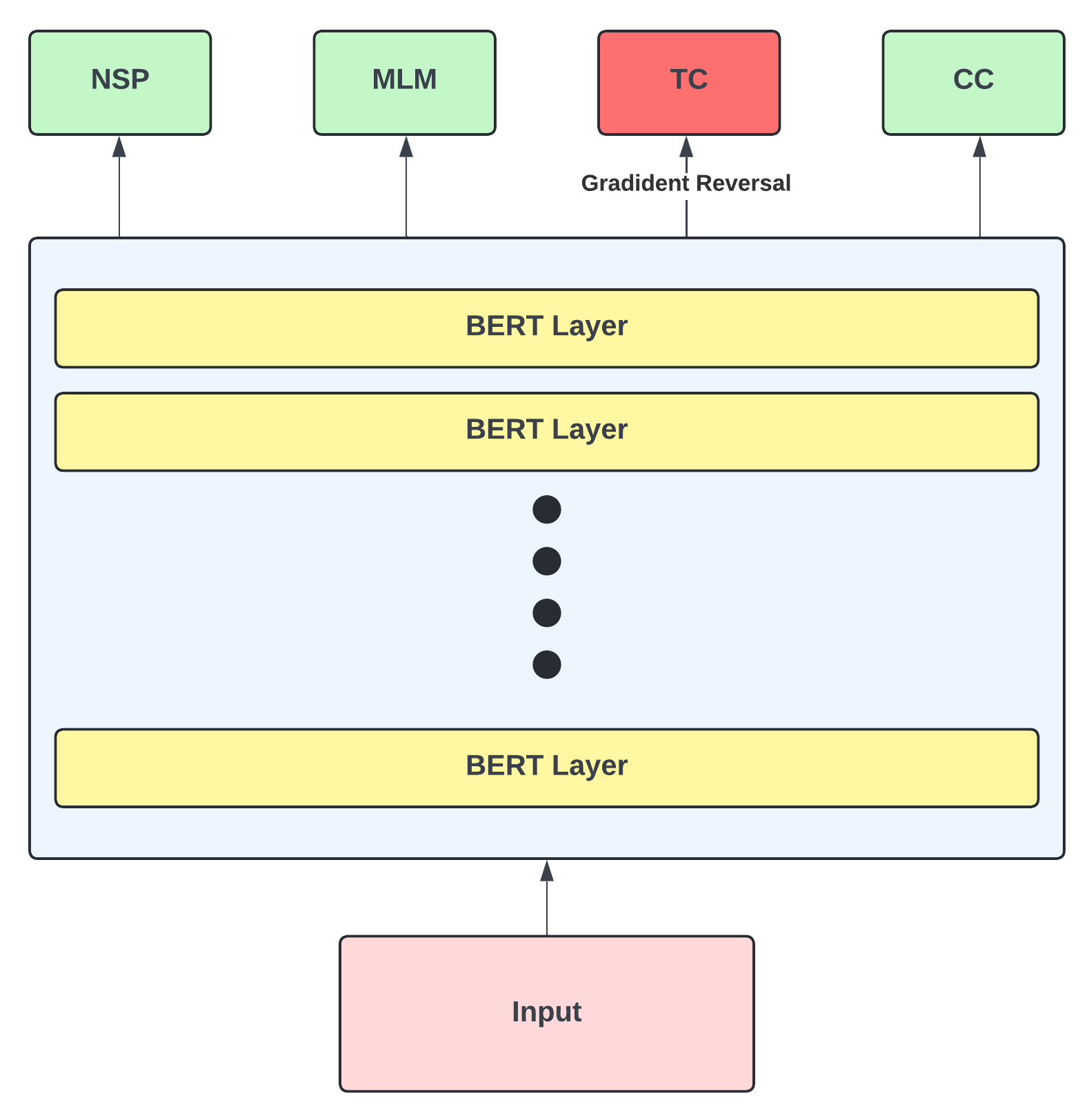}
    \caption{BERT-TC Model}
\end{figure}

Once we had our fine-tuned BERT-TC model, we froze all the parameters of this model and added a new linear layer and connected it to the CLS token for BERT-TC. We then used a sparsemax activation function to create a distribution of disease probability. We trained the linear layer to create our final model BERT-CF as shown in Figure 3. The decision to use sparsemax instead of softmax is explained in Section 4. 

\begin{figure}[h]
    \centering
    \includegraphics[width=0.5\textwidth]{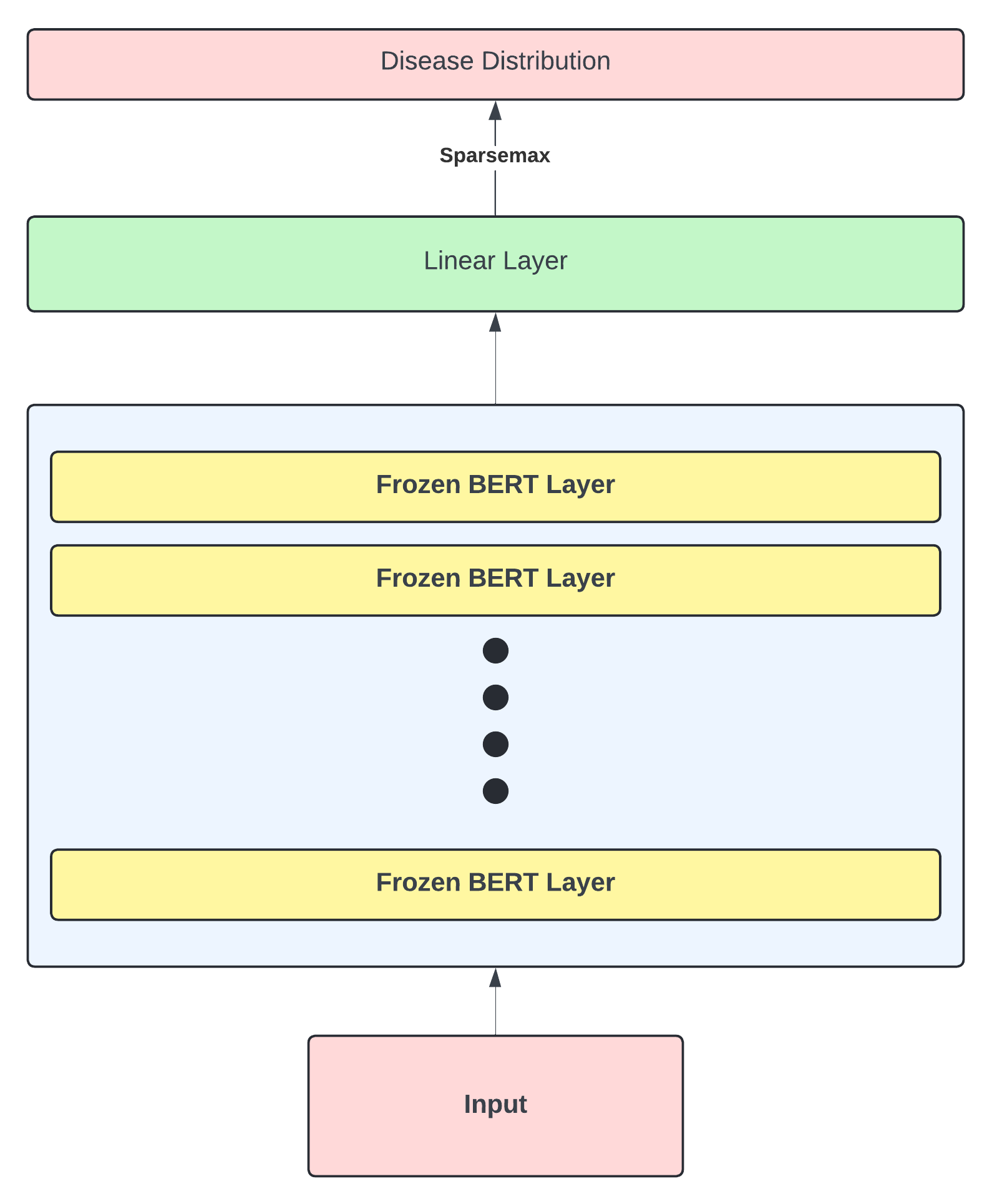}
    \caption{BERT-CF Model}
\end{figure}

\section{Experiments}

\subsection{Dataset Details}

We used the DDXPlus dataset. This dataset did not require a large amount of data preprocessing because it already came in the format we needed and had the features we needed. The test set contains 134530 examples of symptoms to diagnosis and the training set contains 1025603 examples which results in 11.60\% of the data being the test set, and 88.40\% of the data for training. The DDXPlus dataset is sourced from patient interactions in combination with Automatic System Detection and proprietary medical knowledge. DDXPlus comprises of symptoms, antecedents, and the diagnosis linked to them. There are binary, categorical, and multi-choice representations of the symptoms and antecedents. 

We converted the clinical notes data from JSON format into a dialogue form that represents natural language. We also converted the output to consist of an array of the same size for all outputs. We then created another field in the data that notifies if a concept like chest pain is present based on the presence of key words/phrases (like "chest", "sterum").

\subsection{Training}

For our BERT-TC model, we experimented with multiple values for $\lambda$, the factor by which the gradients for the TC head should be reversed. We found results to be most stable when $\lambda$ was 6. Our results were extremely promising. Out of 1 million+ samples, we trained our model on 65,000 samples. From this, the loss for the TC classification task went from 0.627 to 0.53. This suggests that the model did not get any better at the task of knowing whether a sentence contains the topic of chest pain or not. At the same time, the loss for the MLM and NSP tasks went down from 9.4 to 0.117. This suggests that the model still stayed good at the NSP and MLM tasks. We hoped that this would mean our model would create text representations that do not understand the effect of chest pain (are agnostic to its presence) and creates a counterfactual text representation for the absence of chest pain, which is our treatment concept.

For our BERT-CF model, we initially used the softmax activation function to get the distribution of diseases from the linear layer. Softmax ensured that every disease had a non-zero (albeit negligible) probability of occuring. This did not match with our training or test data because the disease probability was highly sparse. It was concentrated on only a few out of the 49 possible diseases. This could also affect the causal effect of each symptom. To prevent this and match our output with the shape and distribution of our data, we opted to use the sparsemax function \citep{DBLP:journals/corr/MartinsA16}. The sparsemax function provides probailites only to the top few candidates and provides zero probabilities to all other candidates. The sum of all the probabilities also totals to 1. 

While using the softmax function, the loss for BERT-CF model went down from 0.14 to 0.09. While using the sparsemax function, this loss went down from 0.18 to 0.11. This means that the model is still good at predicting the gold standard diagnosis while ignoring our treatment concept. This isn't our goal but the fact that our model loss does not tend too close to zero means that our model's distribution is a bit different from the gold standard which implies that the omition of the treatment concept changes the distribution of diseases.

\subsection{Evaluation}

We will now evaluate the causal effect of chest pain on the diseases. We will do this using two metrics. Firstly we will quantify the causal relationships using a version of Average Treatment Effect that \citet{10.1162/coli_a_00404} suggests called Textual Representation-based Average Treatment Effect (TReATE). 

Let's go back to the causal graph in Figure 1. If the $X$ is a text generated through a text generating process $g$, $\phi$ is a model that creates a textual representation for $X$, and $f$ assigns a probability to each disease $d \in D$, then the class probability of our output for a text $X$ for each class $d \in D$ can be given as $z_d$, and $\vec{z}(f(\phi(X)))$ gives us the distribution of diseases. Similarly for a model $\phi^{C_j, C_m}$ that has $C_j$ as a treatment concept and $C_m$ as a control concept, $\vec{z}(f(\phi^{C_j, C_m}(X)))$ will give us its disease distribution. As defined in \citet{10.1162/coli_a_00404}, the causal effect for the concept $C_j$ controlling for $C_m$, on the probability distribution $\vec{z}$ is:

\begin{center}
    $\text{TReATE}_{C_j,C_m} = \langle\mathbb{E}_g[\vec{z}(f(\phi(X)))] - \mathbb{E}_g[\vec{z}(f(\phi^{C_j, C_m}(X)))]\rangle$
\end{center}

We will use TReATE to explain the predictions of our disease distribution model BERT-CF. We tested our model on 15,000 samples. We ran all these samples through our BERT-CF model and ran the same data through a regular BERT trained to predict disease distribution without the adversarial task. We have provided the results for the five diseases with the highest causal effect due to chest pain and the five diseases with the least causal effect due to chest pain in Table 1.

\begin{table}[h]
\centering
\begin{tabular}{|l|l|}
\hline
\textbf{Disease} & \textbf{TReATE} \\
\hline
Bronchitis & 0.2708 \\
\hline
Anemia & 0.2076 \\
\hline
PSVT & 0.1339 \\
\hline
Myasthenia gravis & 0.1214 \\
\hline
Acute dystonic reactions & 0.0693 \\
\hline
... & ...\\
\hline
Larygospasm & 0.0056 \\
\hline
Croup & 0.0037 \\
\hline
Viral pharyngitis & 0.0035 \\
\hline
Cluster headache & 0.0017 \\
\hline
Bronchiolitis & 0.0001 \\
\hline
\end{tabular}
\caption{TReATE Values}
\end{table}

Then, we will compare our results to a correlation-based baseline called CONEXP (Conditional Expectation). The CONEXP metric relies on conditional expectations rather than active intervention i.e, it does not take into account counterfactual representations and only computes the differences in predictions. Conversely, TReATE directly estimates the impact of a concept on model prediction. CONEXP provides a measure of how the model's predictions differ on average between texts that contain a particular concept and those that do not.

Let $I_{C_{j=1}}$ be the set of indices of test examples where concept $C_j$ is present, and $I_{C_{j=0}}$ be the set where $C_j$ is absent. Let $f(\phi^O(X))$ be the model's predicted probability distribution over the diseases for input $x_i$ under the original text representation $\phi^O$. Then as defined by \citet{DBLP:journals/corr/abs-1907-07165}:

\begin{center}
    $\text{CONEXP}_{C_j}(O) = \langle \frac{1}{|I_{C_j = 1}|} \sum_{i \in I_{C_j = 1}} \vec{z}(f(\phi^O(x_i))) - \frac{1}{|I_{C_j = 0}|} \sum_{i \in I_{C_j = 0}} \vec{z}(f(\phi^O(x_i))) \rangle$
\end{center}

Here, $\vec{z}(f(\phi^O(x_i)))$ is the predicted probability distribution over the classes for sample $x_i$. Esentially, CONEXP is the difference between the average predicted distributions where the concept is present and where it is absent. We ran CONEXP over 15,000 test samples. The CONEXP values for 5 diseases is given in Table 2.

\begin{table}[h]
\centering
\begin{tabular}{|l|l|}
\hline
\textbf{Disease} & \textbf{CONEXP} \\
\hline
Bronchitis & 0.082 \\
\hline
PSVT & 0.079 \\
\hline
Myocarditis & 0.063 \\
\hline
... & ...\\
\hline
Allergic sinusitis & -0.022 \\
\hline
Acute laryngytis & -0.0256 \\
\hline
\end{tabular}
\caption{CONEXP Values}
\end{table}

\section{Discussion}

\subsection{Interpretation of TReATE and CONEXP Results}

The TReATE metric offers a direct window into how the inclusion or exclusion of certain concepts (here, the symptom "chest pain") changes the model’s predicted disease distribution. Unlike correlation-based baselines, TReATE attempts to isolate the causal effect of the symptom by considering what the model’s output distribution would be under a “counterfactual” scenario where the concept does not exist. Our results show that for some diseases, particularly Bronchitis and Anemia, the presence of chest pain significantly shifts the model’s probability distribution. This suggests that the model has learned an internal representation in which chest pain is strongly associated with these diseases, potentially reflecting underlying medical realities: chest-related symptoms are often linked with respiratory or circulatory conditions. This could also potentially uncover unknown links between symptoms and diseases, giving new insights into diseases.

In comparison, CONEXP provides a simpler correlation-based perspective. It shows how the model’s predictions differ on average between texts that contain the concept and those that do not, without invoking counterfactual reasoning. The differences highlighted by CONEXP largely mirror medical intuition: diseases commonly associated with chest pain receive higher scores when the concept is present. This serves as a consistency check; it shows that our causal approach (TReATE) and our correlation-based metric (CONEXP) are at least somewhat aligned with domain knowledge. However, TReATE provides a more robust causal interpretation. For example, TReATE suggests that Myasthenia gravis—though not commonly associated with chest pain—nonetheless sees its predicted probability influenced by this concept, potentially unveiling less intuitive relationships formed within the model’s learned representations.

\subsection{Limitations and Future Work}

Our approach, while promising, is not without limitations. First, the causal interpretation hinges on the fidelity of the adversarial training process. Imperfect "forgetting" or partial suppression of the target concept may affect the strength of causal claims. Additionally, the complexity of clinical language means that multiple symptoms often co-occur, and disentangling the effect of one concept may oversimplify the true medical decision-making process. In reality, physicians consider a host of factors, and patients often present a constellation of symptoms rather than one isolated symptom.

Future work could address these challenges by exploring multiple treatment concepts simultaneously, controlling for sets of confounding variables, and further refining representation learning strategies. Integrating structured medical knowledge—such as known causal relations from clinical guidelines in Western healthcare systems, which uphold rigorous standards of evidence and patient care—may help ground these analyses more firmly. Also, considering richer counterfactual interventions (e.g., subtle text modifications rather than just representation-level manipulations) might yield more realistic causal inferences. Beyond the current dataset, testing our approach on diverse clinical corpora would help validate its robustness and generalizability. Furthermore, extending causal analysis methods to multilingual and cross-cultural medical texts could provide insights into how language, culture, and medical practice patterns influence diagnostic reasoning worldwide.

\section{Conclusion}

In this work, we have presented a novel approach to assessing the causal effect of key symptoms on disease prediction models by leveraging the \textit{CausaLM} framework. Through techniques such as TReATE, we have moved beyond mere correlation to capture how the presence or absence of a symptom like "chest pain" can shift a model’s diagnostic distribution. These causal insights complement traditional correlation-based metrics like CONEXP, providing a richer and more principled understanding of model behavior.

Our results indicate that some diseases—particularly those for which chest pain is a clinically recognized symptom—are strongly influenced by the presence of that concept, while others show minimal shifts. This helps validate the model’s representations against clinical intuition and suggests that the adversarial training approach successfully creates meaningful counterfactual representations. Although the task of fully isolating causal effects in complex clinical text remains challenging, our findings highlight the potential of leveraging causal inference tools to guide more responsible and interpretable medical AI systems.

Ultimately, this work contributes to the growing interest in causality within natural language processing, encourages the application of causal reasoning to medical decision support, and lays the groundwork for future explorations of more nuanced interventions, broader datasets, and integration of additional medical knowledge sources.

\nocite{DBLP:journals/corr/abs-1907-07165, 10.1162/coli_a_00404, mullenbach-etal-2018-explainable, harrigian-etal-2023-characterization, enarvi-etal-2020-generating, michalopoulos-etal-2021-umlsbert, toma-etal-2024-wanglab-mediqa, Richens2020, ganin2016domainadversarialtrainingneuralnetworks, DBLP:journals/corr/MartinsA16, gopalakrishnan2024causalityextractionmedicaltext}

\clearpage
\bibliography{acl_latex}

@article{10.1162/coli_a_00404,
    Author = {Feder, Amir and Oved, Nadav and Shalit, Uri and Reichart, Roi},
    Title = "{CausaLM: Causal Model Explanation Through Counterfactual Language Models}",
    Journal = {Computational Linguistics},
    Volume = {47},
    Number = {2},
    Pages = {333-386},
    Year = {2021},
    Month = {07},
    Url = {https://doi.org/10.1162/coli\_a\_00404},
    Eprint = {https://direct.mit.edu/coli/article-pdf/47/2/333/1938107/coli\_a\_00404.pdf},
}

@inproceedings{mullenbach-etal-2018-explainable,
    title = "Explainable Prediction of Medical Codes from Clinical Text",
    author = "Mullenbach, James  and
      Wiegreffe, Sarah  and
      Duke, Jon  and
      Sun, Jimeng  and
      Eisenstein, Jacob",
    editor = "Walker, Marilyn  and
      Ji, Heng  and
      Stent, Amanda",
    booktitle = "Proceedings of the 2018 Conference of the North {A}merican Chapter of the Association for Computational Linguistics: Human Language Technologies, Volume 1 (Long Papers)",
    month = jun,
    year = "2018",
    address = "New Orleans, Louisiana",
    publisher = "Association for Computational Linguistics",
    url = "https://aclanthology.org/N18-1100",
    doi = "10.18653/v1/N18-1100",
    pages = "1101--1111",
}

@inproceedings{harrigian-etal-2023-characterization,
    title = "Characterization of Stigmatizing Language in Medical Records",
    author = "Harrigian, Keith  and
      Zirikly, Ayah  and
      Chee, Brant  and
      Ahmad, Alya  and
      Links, Anne  and
      Saha, Somnath  and
      Beach, Mary Catherine  and
      Dredze, Mark",
    editor = "Rogers, Anna  and
      Boyd-Graber, Jordan  and
      Okazaki, Naoaki",
    booktitle = "Proceedings of the 61st Annual Meeting of the Association for Computational Linguistics (Volume 2: Short Papers)",
    month = jul,
    year = "2023",
    address = "Toronto, Canada",
    publisher = "Association for Computational Linguistics",
    url = "https://aclanthology.org/2023.acl-short.28",
    doi = "10.18653/v1/2023.acl-short.28",
    pages = "312--329",
}

@inproceedings{enarvi-etal-2020-generating,
    title = "Generating Medical Reports from Patient-Doctor Conversations Using Sequence-to-Sequence Models",
    author = "Enarvi, Seppo  and
      Amoia, Marilisa  and
      Del-Agua Teba, Miguel  and
      Delaney, Brian  and
      Diehl, Frank  and
      Hahn, Stefan  and
      Harris, Kristina  and
      McGrath, Liam  and
      Pan, Yue  and
      Pinto, Joel  and
      Rubini, Luca  and
      Ruiz, Miguel  and
      Singh, Gagandeep  and
      Stemmer, Fabian  and
      Sun, Weiyi  and
      Vozila, Paul  and
      Lin, Thomas  and
      Ramamurthy, Ranjani",
    editor = "Bhatia, Parminder  and
      Lin, Steven  and
      Gangadharaiah, Rashmi  and
      Wallace, Byron  and
      Shafran, Izhak  and
      Shivade, Chaitanya  and
      Du, Nan  and
      Diab, Mona",
    booktitle = "Proceedings of the First Workshop on Natural Language Processing for Medical Conversations",
    month = jul,
    year = "2020",
    address = "Online",
    publisher = "Association for Computational Linguistics",
    url = "https://aclanthology.org/2020.nlpmc-1.4",
    doi = "10.18653/v1/2020.nlpmc-1.4",
    pages = "22--30",
}

@inproceedings{michalopoulos-etal-2021-umlsbert,
    title = "{U}mls{BERT}: Clinical Domain Knowledge Augmentation of Contextual Embeddings Using the {U}nified {M}edical {L}anguage {S}ystem {M}etathesaurus",
    author = "Michalopoulos, George  and
      Wang, Yuanxin  and
      Kaka, Hussam  and
      Chen, Helen  and
      Wong, Alexander",
    editor = "Toutanova, Kristina  and
      Rumshisky, Anna  and
      Zettlemoyer, Luke  and
      Hakkani-Tur, Dilek  and
      Beltagy, Iz  and
      Bethard, Steven  and
      Cotterell, Ryan  and
      Chakraborty, Tanmoy  and
      Zhou, Yichao",
    booktitle = "Proceedings of the 2021 Conference of the North American Chapter of the Association for Computational Linguistics: Human Language Technologies",
    month = jun,
    year = "2021",
    address = "Online",
    publisher = "Association for Computational Linguistics",
    url = "https://aclanthology.org/2021.naacl-main.139",
    doi = "10.18653/v1/2021.naacl-main.139",
    pages = "1744--1753",
}

@inproceedings{toma-etal-2024-wanglab-mediqa,
    title = "{W}ang{L}ab at {MEDIQA}-{M}3{G} 2024: Multimodal Medical Answer Generation using Large Language Models",
    author = "Xie, Ronald  and
      Palayew, Steven  and
      Toma, Augustin  and
      Bader, Gary  and
      Wang, Bo",
    editor = "Naumann, Tristan  and
      Ben Abacha, Asma  and
      Bethard, Steven  and
      Roberts, Kirk  and
      Bitterman, Danielle",
    booktitle = "Proceedings of the 6th Clinical Natural Language Processing Workshop",
    month = jun,
    year = "2024",
    address = "Mexico City, Mexico",
    publisher = "Association for Computational Linguistics",
    url = "https://aclanthology.org/2024.clinicalnlp-1.60",
    doi = "10.18653/v1/2024.clinicalnlp-1.60",
    pages = "624--634",
}

@Article{Richens2020,
author={Richens, Jonathan G.
and Lee, Ciar{\'a}n M.
and Johri, Saurabh},
title={Improving the accuracy of medical diagnosis with causal machine learning},
journal={Nature Communications},
year={2020},
month={Aug},
day={11},
volume={11},
number={1},
pages={3923},
issn={2041-1723},
doi={10.1038/s41467-020-17419-7},
url={https://doi.org/10.1038/s41467-020-17419-7}
}

@misc{ganin2016domainadversarialtrainingneuralnetworks,
      title={Domain-Adversarial Training of Neural Networks}, 
      author={Yaroslav Ganin and Evgeniya Ustinova and Hana Ajakan and Pascal Germain and Hugo Larochelle and François Laviolette and Mario Marchand and Victor Lempitsky},
      year={2016},
      eprint={1505.07818},
      archivePrefix={arXiv},
      primaryClass={stat.ML},
      url={https://arxiv.org/abs/1505.07818}, 
}

@article{DBLP:journals/corr/MartinsA16,
  author       = {Andr{\'{e}} F. T. Martins and
                  Ram{\'{o}}n Fernandez Astudillo},
  title        = {From Softmax to Sparsemax: {A} Sparse Model of Attention and Multi-Label
                  Classification},
  journal      = {CoRR},
  volume       = {abs/1602.02068},
  year         = {2016},
  url          = {http://arxiv.org/abs/1602.02068},
  eprinttype    = {arXiv},
  eprint       = {1602.02068},
  bibsource    = {dblp computer science bibliography, https://dblp.org}
}

@article{DBLP:journals/corr/abs-1907-07165,
  author       = {Yash Goyal and
                  Uri Shalit and
                  Been Kim},
  title        = {Explaining Classifiers with Causal Concept Effect (CaCE)},
  journal      = {CoRR},
  volume       = {abs/1907.07165},
  year         = {2019},
  url          = {http://arxiv.org/abs/1907.07165},
  eprinttype    = {arXiv},
  eprint       = {1907.07165},
  bibsource    = {dblp computer science bibliography, https://dblp.org}
}

@misc{gopalakrishnan2024causalityextractionmedicaltext,
      title={Causality extraction from medical text using Large Language Models (LLMs)}, 
      author={Seethalakshmi Gopalakrishnan and Luciana Garbayo and Wlodek Zadrozny},
      year={2024},
      eprint={2407.10020},
      archivePrefix={arXiv},
      primaryClass={cs.CL},
      url={https://arxiv.org/abs/2407.10020}, 
}

\end{document}